\documentclass{article}
\usepackage{ijcai24}

\usepackage{times}
\usepackage{soul}
\usepackage{url}
\usepackage[hidelinks]{hyperref}
\usepackage[utf8]{inputenc}
\usepackage[small]{caption}
\usepackage{graphicx}
\usepackage{amsmath}
\usepackage{amsthm}
\usepackage{booktabs}
\usepackage{algorithm}
\usepackage{algorithmic}
\usepackage[switch]{lineno}

\usepackage{threeparttable}
\usepackage{subfig}
\usepackage{mathtools}

\usepackage{amsfonts}
\usepackage{amssymb}

\usepackage{arydshln}

\title{Enhancing Sustainable Urban Mobility Prediction with Telecom Data: \\ A Spatio-Temporal Framework Approach}

\author{
ChungYi Lin$^{1,2}$
\and
Shen-Lung Tung$^1$\and
Hung-Ting Su$^{2}$\And
Winston H. Hsu$^{2,3}$\\
\affiliations
$^1$Internet of Things Laboratory, Chunghwa Telecom Laboratories\\
$^2$National Taiwan University\\
$^3$Mobile Drive Technology\\
}

\begin{document}

\maketitle

\begin{abstract}
Traditional traffic prediction, limited by the scope of sensor data, falls short in comprehensive traffic management. Mobile networks offer a promising alternative using network activity counts, but these lack crucial directionality. Thus, we present the TeltoMob dataset, featuring undirected telecom counts and corresponding directional flows, to predict directional mobility flows on roadways. To address this, we propose a two-stage spatio-temporal graph neural network (STGNN) framework. The first stage uses a pre-trained STGNN to process telecom data, while the second stage integrates directional and geographic insights for accurate prediction. Our experiments demonstrate the framework's compatibility with various STGNN models and confirm its effectiveness. We also show how to incorporate the framework into real-world transportation systems, enhancing sustainable urban mobility.
\end{abstract}

\section{Introduction}
Effective traffic management is crucial for intelligent transportation systems \cite{xie2020urban,lv2021deep}. Traditional methods rely on costly detectors with limited coverage \cite{sen2012kyun,li2018diffusion,guo2019attention}. With over 71\% of the global population connected to mobile networks \cite{cisco2021cisco}, cellular traffic activities \cite{jiang2022cellular} offer valuable insights. The \textit{count of cellular traffic} (i.e., \textbf{cellular traffic flow}) can proxy traffic conditions \cite{lin2021multivariate}. However, the \textbf{lack of directionality} in cellular traffic flows from road areas limits understanding commuting patterns and easing congestion, thus reducing their utility.

\begin{figure}[ht]
\centering
\includegraphics[width=1\linewidth]{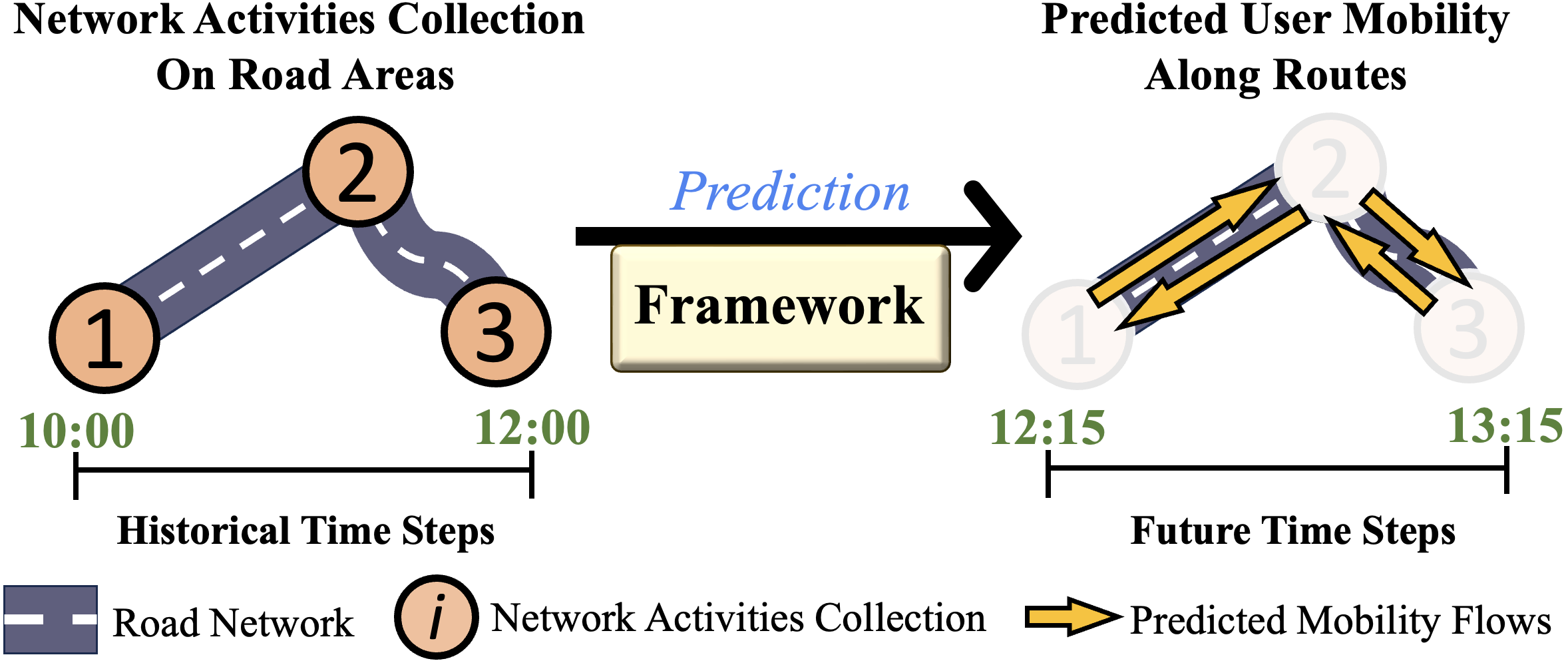}
\caption{Overview of the task and framework. Network activities collected at road areas (points 1 to 3) act as proxies for traffic conditions but lack crucial directionality for accurate traffic management. Our framework leverages non-directional telecom data from past time steps to predict future directional mobility flows, enhancing its utility for urban computing.}
\label{fig_overview}
\end{figure}

To extract directionality for traffic management, we present a task utilizing cellular traffic flows from selected road areas to predict \textit{user mobility counts} (i.e., \textbf{mobility flows}) along routes (as Figure \ref{fig_overview}). This enhances the utility of undirected telecom data by providing directional insights, reducing costs and environmental impact associated with sensor deployment, and aligning with the Sustainable Development Goals (SDG)\footnote{https://sdgs.un.org/goals/goal11} for urban sustainability. To support this task, we propose the Tel-to-Mob dataset, including undirected telecom-based flows from 34 roads and directed mobility flows for 84 routes, with analysis to exhibit its relevance to road structure.

We identified two main challenges: \textit{Magnitude Disparity}, where cellular traffic flows capture all users in an area, unlike mobility flows that reflect specific directional movements; and \textit{Amount Disparity}, where a single road area being part of multiple routes leads to misalignments, hindering direct mapping from cellular traffic to mobility flows, a gap not addressed by current models (e.g., \cite{li2023dynamic,lin2024teltrans}). To tackle these, we propose a \textbf{Spatio-Temporal Graph Neural Network (STGNN) Framework} with two stages. \textit{Stage 1} employs a pre-trained STGNN to extract features from cellular traffic flows. \textit{Stage 2} transforms these features to integrate directionality and enhances them with geographical insights, using another STGNN to capture spatio-temporal dynamics and predict future mobility flows.

Overall, our main contributions:

$\bullet$ \textbf{What Addressed}: We use telecom-based flows to forecast directional mobility flows, overcoming traffic sensor limitations and advancing sustainable urban living.

$\bullet$ \textbf{Who Involved}: We use anonymous data from extensive mobile users provided by a cooperating telecom operator.

$\bullet$ \textbf{How Evaluated}: Our framework's effectiveness is evaluated based on prediction accuracy. All related data and code are accessible at: https://github.com/cy07gn/TeltoMob.

\clearpage

\section{TeltoMob dataset}
\noindent \textbf{Preliminary.} As cellular traffic is collected from mobile users moving across adjacent areas \cite{zhang2018citywide}, it exhibits spatial correlations \cite{wang2018spatio,wang2022spatial}. However, the primary goal usually focuses on enhancing network resource allocation in specific areas \cite{yao2021mvstgn,zhao2021spatial} or at base stations \cite{wang2018spatio}, as well as inducing energy savings \cite{lin2021data} and improving resource scheduling \cite{he2020semi}.

However, as we aim to utilize cellular traffic for transportation evaluation, the lack of directionality reduces its practicality for traffic management. Thus, we introduce the \textbf{TeltoMob} dataset, which contains undirected telecom-based flow and directed mobility flow among road sections.

\subsection{Definitions}
\noindent $\bullet$  {\textbf{Geographical Cellular Traffic (GCT)}}: A cellular traffic record with its originating GPS coordinates, as Table \ref{tab:gct}A.

\begin{figure}[ht]
\centering
\includegraphics[width=0.95\linewidth]{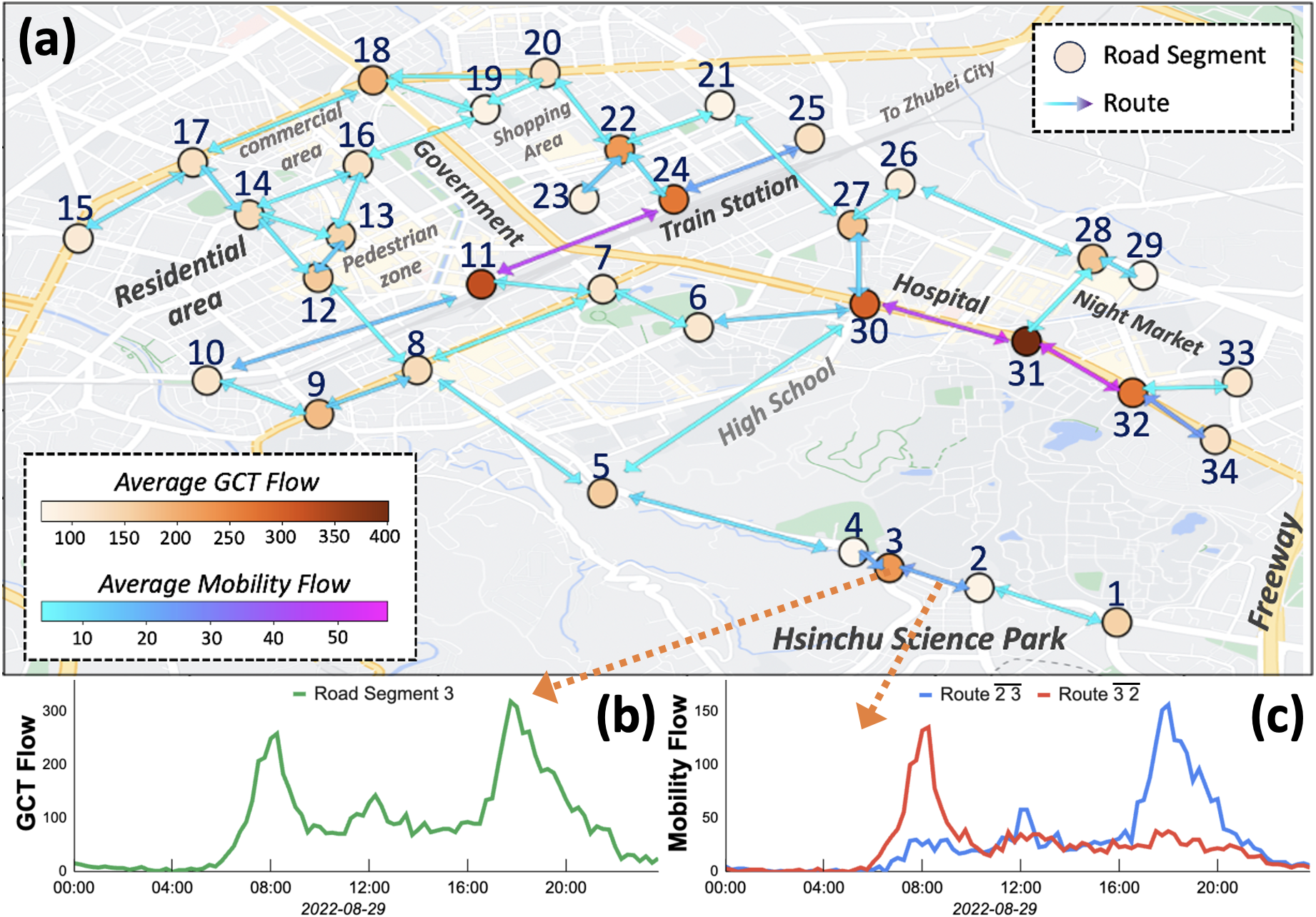}
\caption{Overview of data collected from 34 road segments, including 84 directional routes in Hsinchu City. (a) The map depicts GCTs sourced from user activity, while mobility (i.e., GCT pairing) is determined by associating GCT records appearing in adjacent segments along routes. Color intensity represents the average volumes of GCT and mobility flows. (b) Sample daily GCT flow pattern. (c) Sample daily mobility flow pattern.}
\label{fig_road_segments}
\end{figure}

\noindent $\bullet$  {\textbf{Road Segment}}: A 20m x 20m road area, based on the average road size in the Proof-of-Concept area, as Figure \ref{fig_road_segments}(a).

\noindent $\bullet$ {\textbf{Route}}: A directional pathway from start road segment $i$ to end segment $j$, denoted as \textbf{$\overline{ij}$}.

\noindent $\bullet$ {\textbf{GCT Flow}}: The count of GCT records on a road segment, accumulated over a fixed time interval, as Figure \ref{fig_road_segments}(b).

\noindent $\bullet$ {\textbf{GCT Pairing}}: An entry by associating two GCT records from consecutive segments of the same user, as Table \ref{tab:gct}B.

\noindent $\bullet$  {\textbf{Mobility Flow}}: 
The count of \textit{GCT pairings} along routes, recorded over fixed time intervals (see Figure \ref{fig_road_segments}(c)), offers an alternative to physical detectors, aligning with SDG aims.

\begin{table}[ht]
  \centering
  \footnotesize

  \begin{tabular*}{\linewidth}{@{\extracolsep{\fill}} llll}
    \toprule
    \multicolumn{4}{c}{\textbf{A. Examples of GCT Records}} \\ 
    \midrule
    {IMEI$^{1}$} & {Date} & {Latitude} & {Longitude} \\
    \midrule
    ... \\
    524edbbd5122 & 2022-09-06 18:02:17 & 24.801066 & 120.987103 \\ 
    a63cc2cc752e & 2022-09-06 18:02:23 & 24.801219 & 120.987091 \\ 
    f4f79deaff0c & 2022-09-06 18:02:30 & 24.801246 & 120.987090 \\ 
    ... \\
    \bottomrule
  \end{tabular*}

  \vspace{2mm} 

  \begin{tabular*}{\linewidth}{@{\extracolsep{\fill}} lll}
    \toprule
    \multicolumn{3}{c}{\textbf{B. Examples of GCT Pairings}} \\ 
    \midrule
    {Pairing$^{2}$} & {Start Time} & {End Time} \\
    \midrule
    ... \\
    524edbbd5122 & 2022/09/06 18:02:17 & 2022/09/06 18:04:02 \\
    a63cc2cc752e & 2022/09/06 18:02:23 & 2022/09/06 18:11:52 \\
    f4f79deaff0c & 2022/09/06 18:02:30 & 2022/09/06 18:11:51 \\
    ... \\
    \bottomrule
  \end{tabular*}
   \scriptsize
  \begin{tablenotes}[flushleft]
    \item [$^{1}$] $^{1}$IMEI, or International Mobile Equipment Identity, is hashed for user privacy.
    
    \item [$^{2}$] $^{2}$ Pairings indicate mobility along a route.
  \end{tablenotes}

  \caption{Examples of GCT Records and GCT Pairings.}
  \label{tab:gct}
  
\end{table}

\subsection{Data Collection and Processing}

\noindent \textbf{Location Selection:}

$\bullet$  {\textit{Road Segment}}. In collaboration with City Authorities, we selected 34 road segments based on \textit{{criteria}} like daily commutes, and congestion-prone areas. The segments are near areas with distinct environments, including universities, shopping centers, and science parks.

$\bullet$  {\textit{Route}}. After identifying road segments, we determined 84 directional routes based on the road network structure, facilitating GCT record pairing to capture mobility. Each route connects a start and end road segment.

\noindent \textbf{Data Sourcing:}

$\bullet$  {\textit{GCT Records}}. All GCT records are stored in the telecom company's Geographical Cellular Traffic Database. We extracted GCTs from 34 road segments, focusing on essential data fields—IMEI, recording time, and coordinates—for time efficiency, as shown in Table \ref{tab:gct}A.

$\bullet$  {\textit{GCT Pairings}}. We match two GCT records with the same IMEI number (i.e., the same user) across adjacent road segments, originating from the start and end road segments, respectively. The time difference between these records is kept within a 15-minute window to exclude pedestrian or non-vehicular traffic, thus focusing on vehicle movements. Table \ref{tab:gct} displays the pairing results for route $\overline{30 \, 31}$.

\noindent \textbf{Processing:} \textit{GCT} and \textit{mobility flows} denote the cumulative counts of \textit{GCT records} and \textit{GCT pairings} at 15-minute intervals, respectively, revealing unique temporal patterns for each road segment and route over time.

\subsection{Data Privacy Protection}
Data privacy is paramount in telecom data. Here's how we protect user anonymity and privacy for our task:

\noindent \textbf{Location Constraints.} We restrict data collection to road segments, avoiding sensitive areas like commercial or residential districts. We focus on GCTs from these segments, preventing tracking of journeys or user pattern identification.

\noindent \textbf{Data Aggregation.} GCT flow is the cumulative count of GCT records that masks individual identities, securing user information for telecom data use.

\noindent \textbf{International Standards.} Our partner telecom company follows ISO 27001 standards, ensuring sensitive data management and access are rigorously controlled.
\subsection{Data Analysis}

\noindent \textbf{Descriptive Statistics.}
Table \ref{tab:statistics} summarizes the descriptive statistics of our dataset from 2022/08/28 to 2022/09/27 with 2,976 samples from 34 road segments and 84 routes. Notably, segment 31 near a hospital has the highest GCT flow with 400.58 entries per 15 minutes, and Route $\overline{30\,31}$, linking downtown to the freeway, records the highest mobility flow with 57.82 movements per 15 minutes.

\begin{table}[ht]
  \centering
  
  \begin{threeparttable}
    \footnotesize 
    \setlength{\tabcolsep}{2pt} 
    \renewcommand{\arraystretch}{1.1} 
    \begin{tabular}{@{}lllllll@{}}
      \toprule
      \textbf{Type} & \textbf{\#Samples} & \textbf{\#Amount} & 
      \textbf{Interval} &
      \textbf{Avg.} & \textbf{STD}\tnote{1} & \textbf{Max Avg.}   \\
      \midrule
      GCT & 2976 & 34 & 15-min. & 159.9 & 116.59 & 400.5   \\
      Mobility & 2976 & 84 & 15-min. & 12.9 & 11.03 & 57.8  \\
      \bottomrule
    \end{tabular}

     \scriptsize
  
    \begin{tablenotes}[flushleft]
      \item[1] Standard deviation.
     
    \end{tablenotes}
  \end{threeparttable}

  \caption{Descriptive Statistics of the Dataset.}
  \label{tab:statistics}
\end{table}

\noindent \textbf{Magnitude Discrepancy.}
Table \ref{tab:statistics} shows that the average GCT flow markedly exceeds that of the mobility flow. This difference is due to the GCT flow including all users—stationary and pedestrians—without considering direction, while mobility flow counts directional movements between segments, typically vehicular. Thus, GCT flow reflects broader user activity, and mobility flow precisely indicates directional vehicular traffic.

\begin{figure}[ht]
\centering
\includegraphics[width=0.96\linewidth]{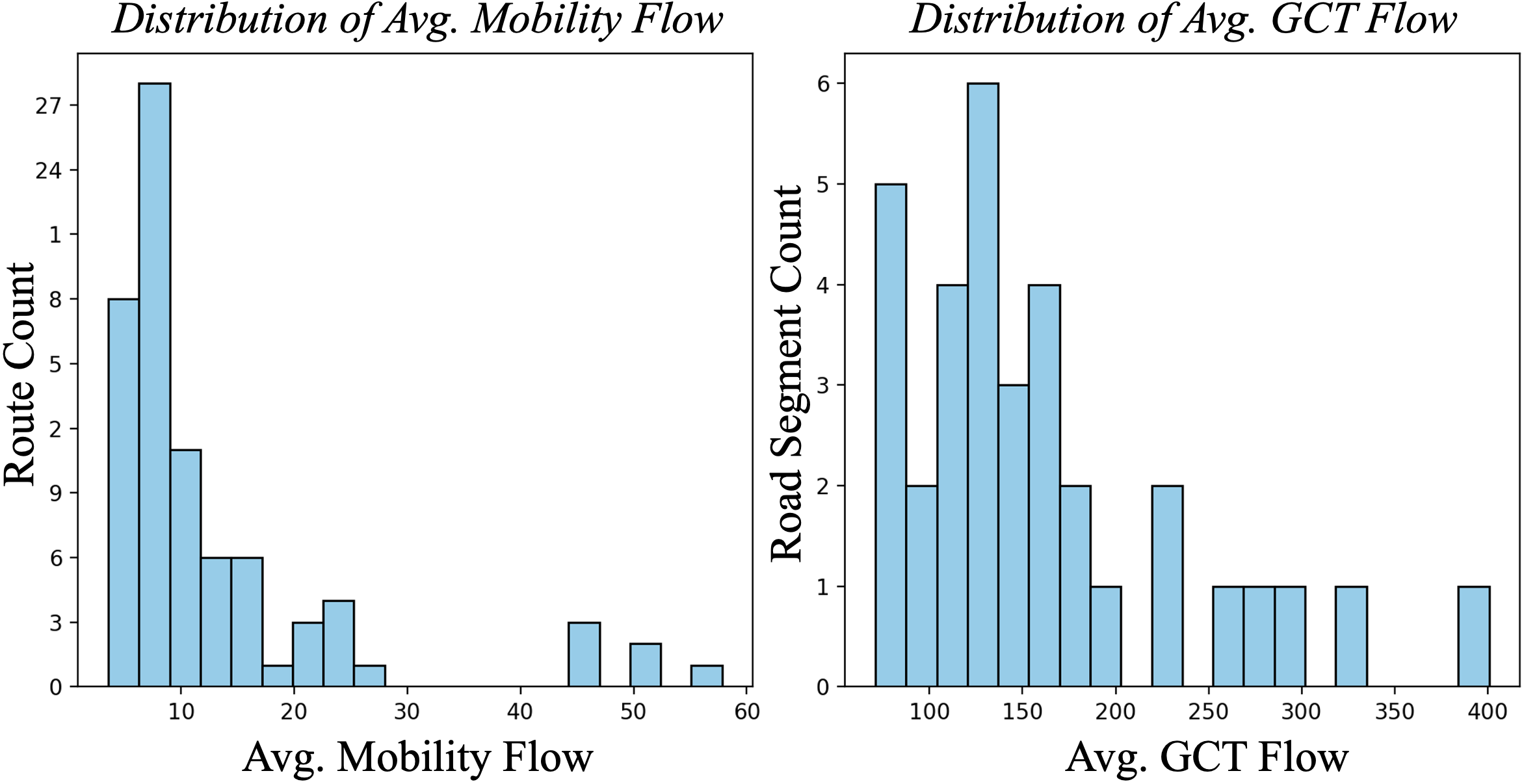}
\caption{
Histograms showing average GCT and mobility flow distribution in our dataset. The x-axis indicates flow intervals, and the y-axis counts road segments and routes. The right-skewed distribution highlights low traffic on most routes, with a few experiencing high volumes, typical of urban road network hierarchies.
}
\label{Histogram}
\end{figure}

\noindent \textbf{Flow Distribution Analysis.}
Figure \ref{Histogram} shows the distribution of average GCT and mobility flows for road segments and routes. The right skew indicates low traffic in most locations, with few experiencing high volumes. This reflects the typical urban network structure \cite{peng2016computational,babu2020toward} where arterials carry main traffic, while local streets have less flow. The dataset accurately reflects real-world traffic trends, proving valuable for urban.

\begin{figure} [ht]
    \centering
  \subfloat[\label{temporal_5to4}]{%
        \includegraphics[width=0.53\linewidth]{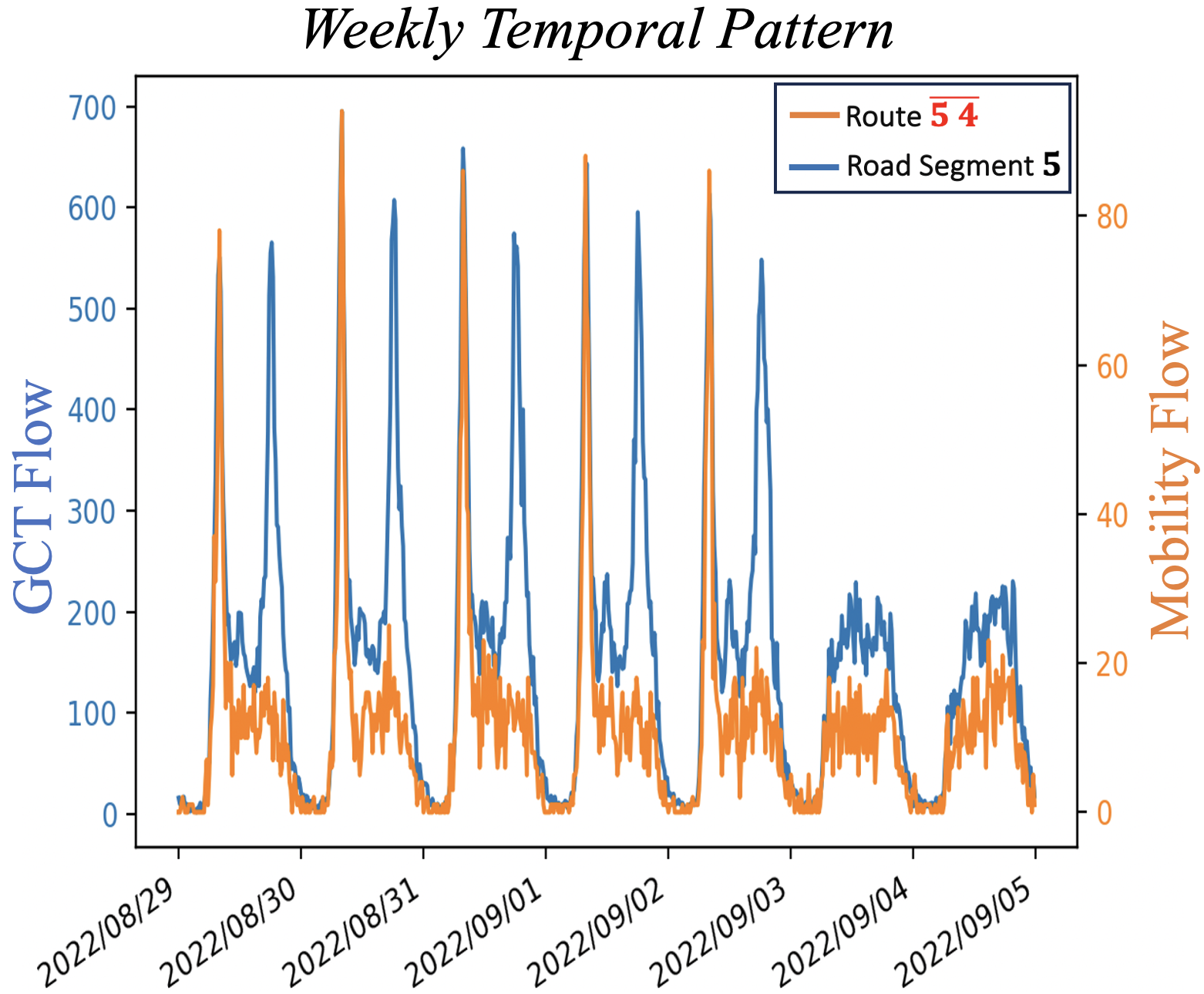}}
  \hfill
  \subfloat[\label{scatter_5to4}]{%
        \includegraphics[width=0.46\linewidth]{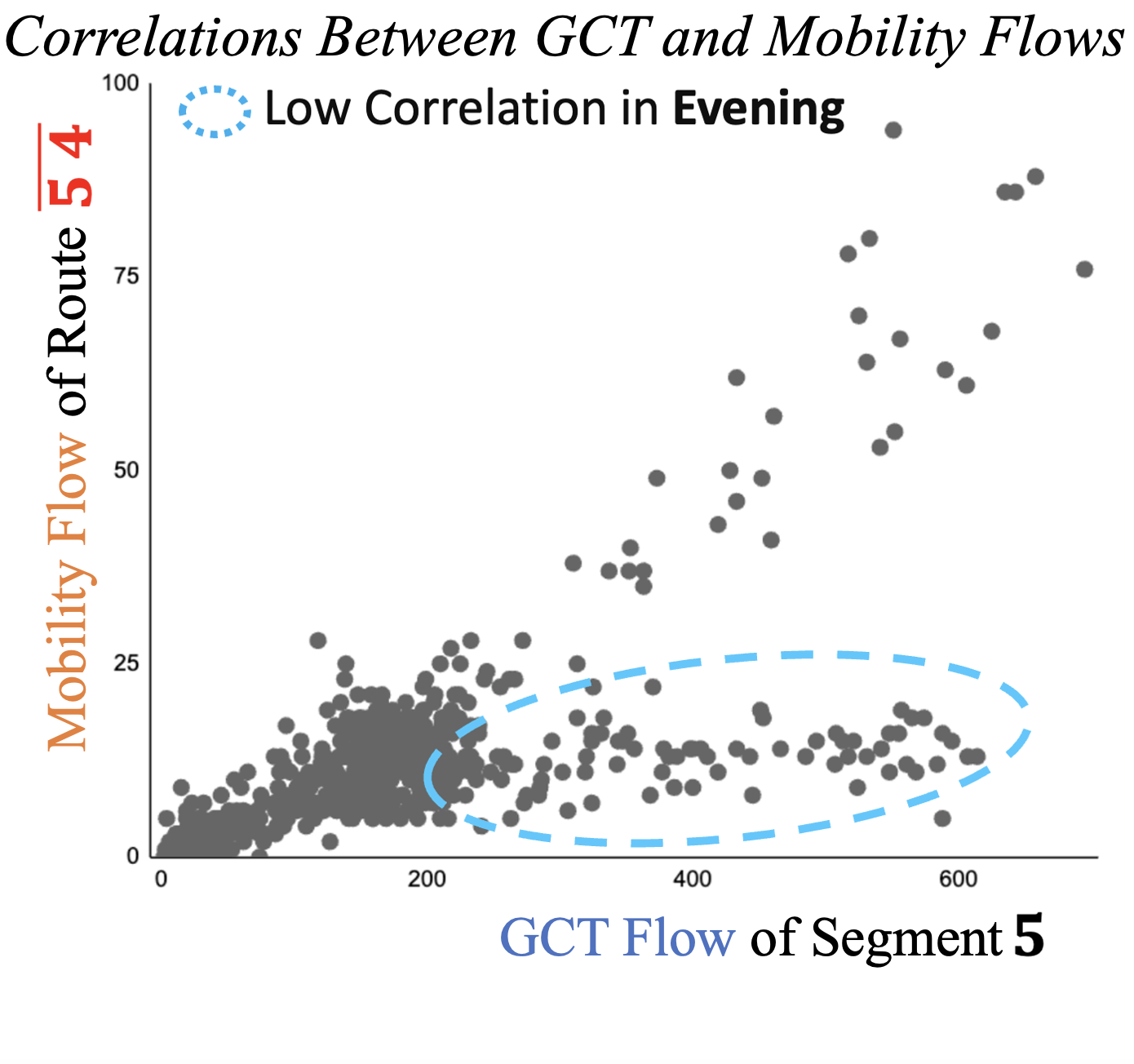}}
   
  \caption{Relationships between GCT and mobility flows. (a) Weekly patterns show morning peaks in mobility flow for work commutes, with less evening traffic, unlike GCT flow, which reflects all directional activities. (b) Although correlations are generally positive, reduced evening mobility compared to persistent high GCT flows leads to lower correlations (indicated by blue ovals).}
  \label{error_rmt_visual} 
\end{figure}

\noindent \textbf{Relationships Between Mobility and GCT Flows.}
We analyzed the correlation between mobility flow on route $\overline{5\,4}$, connecting residential areas to Hsinchu Science Park, and GCT flow on overlapping segment $5$. The weekly trends (Figure \ref{temporal_5to4}) show generally similar patterns, but with some distinct variations during evening commutes. Mobility flow peaks in the morning for work commutes on $\overline{5\,4}$ and declines in the evening as commuters use the opposite route $\overline{4\,5}$. In contrast, GCT flow, which represents all user activities on segment $5$, peaks during both morning and evening commutes, thus diverging from mobility flow. Figure \ref{scatter_5to4} depicts this varying correlation, although typically positive, with some weakening (as blue ovals) due to lower evening mobility on $\overline{5\,4}$ despite high GCT flows. These findings highlight the interplay between these flows, influenced by time and direction.

\begin{figure}[ht]
\centering
\includegraphics[width=0.55\linewidth]{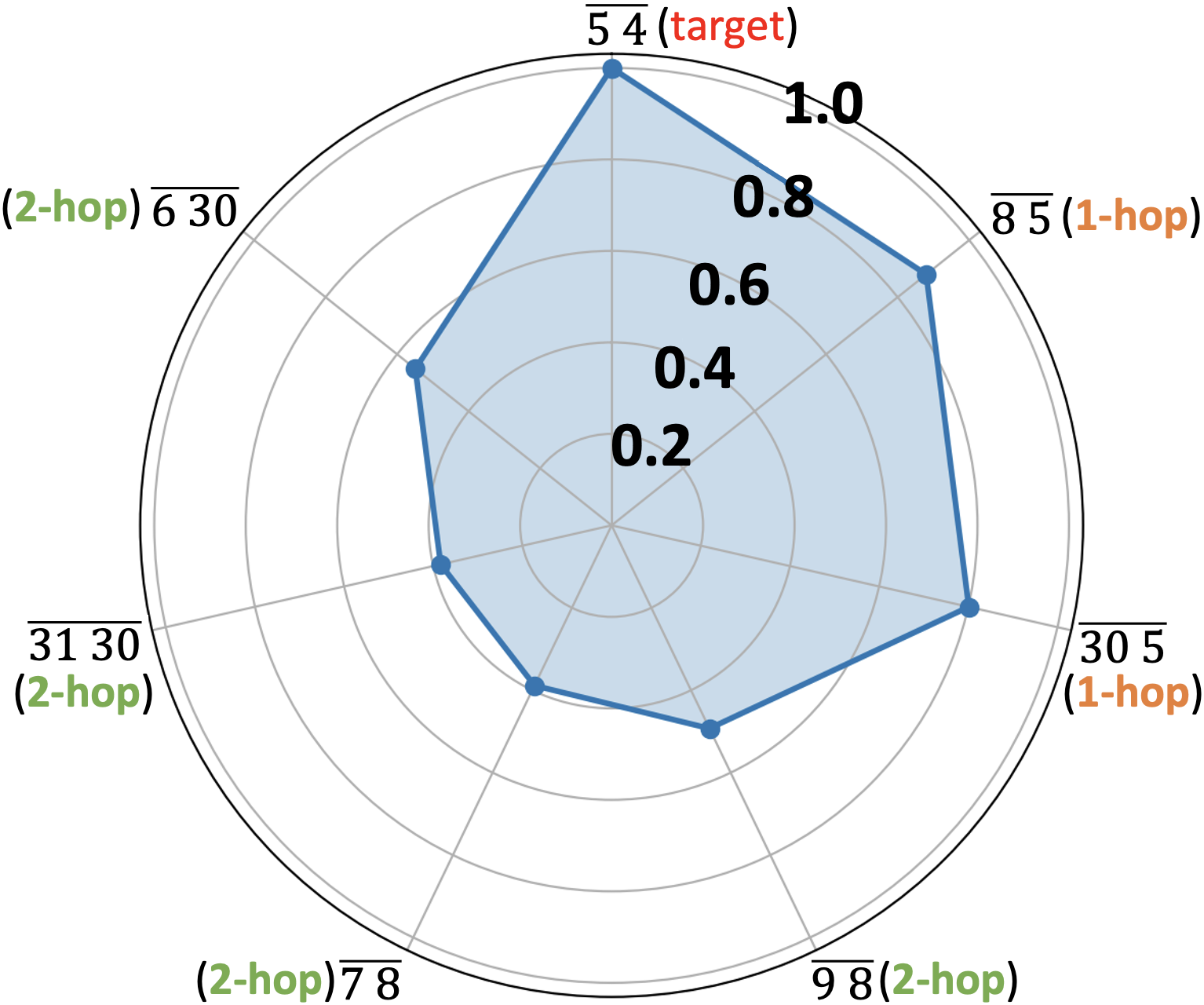}
\caption{Pearson correlation analysis of daily mobility flow for route $\overline{5\,4}$ and its upstream routes on 2022/09/05 reveals strong correlations with 1-hop upstream routes and weaker ones with 2-hop upstream routes, indicating a diminishing impact from distant routes. This finding directs our framework's emphasis on 1-hop upstream route correlations for understanding traffic continuity.}
\label{radar}
\end{figure}

\noindent \textbf{Interactions Between Route's Neighbors.}
Acknowledging that the mobility flow of a route is influenced by upstream movements as users transition from upstream to downstream areas, we analyzed its correlation with upstream neighbors. Focusing on route $\overline{5\,4}$ and its upstream routes on 2022/9/05 as an example, we employed Pearson correlation coefficients \cite{cohen2009pearson} to assess daily patterns. Figure \ref{radar} reveals strong correlations (above 0.8) with direct upstream routes ($\overline{8\,5}$ and $\overline{30\,5}$), signifying their significant impact. Conversely, 2-hop upstream routes exhibit weaker correlations (0.3 to 0.6), suggesting a reduced impact with increased distance. This chart highlights the critical role of the nearest upstream routes in ensuring traffic flow continuity, which informs the geographical insights incorporated in our framework.

\clearpage
\begin{figure*}
    \centering
    \includegraphics[width=0.99\textwidth]{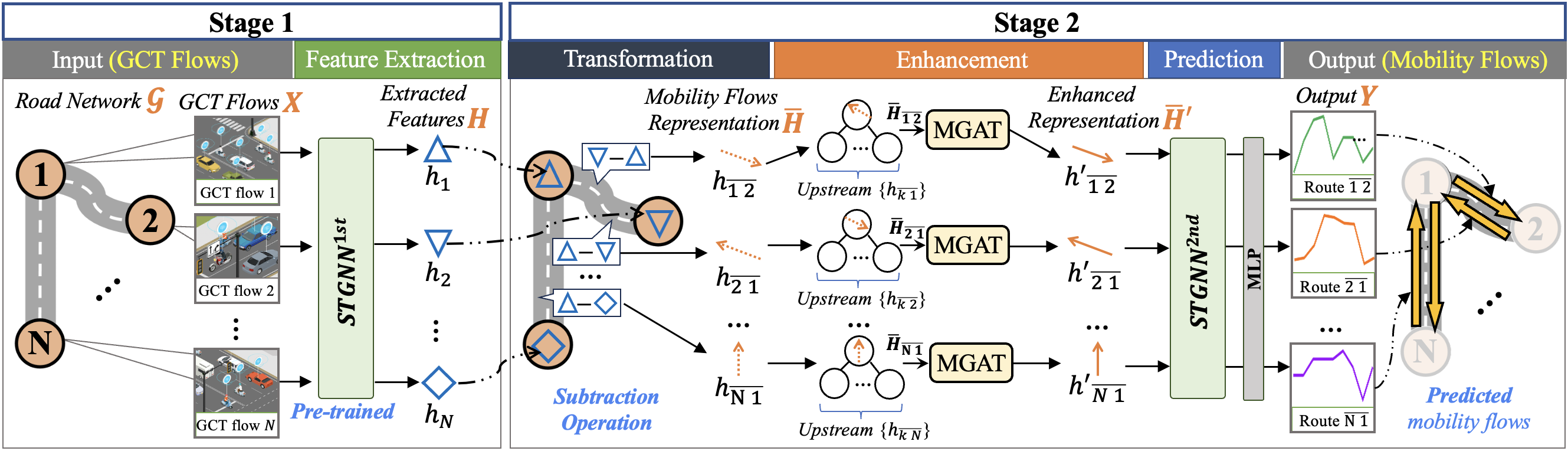}
    \caption{Overview of the proposed two-stage STGNN framework. \textbf{Stage 1} employs a pre-trained STGNN ($STGNN^{1st}$) to extract features from GCT flows (as blue shapes). \textbf{Stage 2} comprises three steps:     \textit{{transformation step}} derives initial representations (as dotted orange arrows) aligned with the amount of mobility flow while incorporating directionality. The \textit{{enhancement step}} integrates each mobility flow representation with its upstream neighbors using the Multi-Channel Graph Attention (MGAT), resulting in enhanced representations (as solid orange arrows). \textit{{prediction step}} utilizes a secondary STGNN ($STGNN^{2nd}$) to generate mobility flow predictions.
    }
    \label{sub_fig:framework}
\end{figure*}
\section{Methodology}

\subsection{Task Definition}
Using undirected \textbf{N} GCT flows from past steps ($T_{in}$) to forecast directional \textbf{M} mobility flows for future steps ($T_{out}$).

\subsection{Framework Overview}
As Figure \ref{sub_fig:framework}, our framework functions in two stages to address the \textbf{\textit{magnitude disparity}} among GCT and mobility flows, and the \textbf{\textit{amount disparity}} among 32 segments and 84 routes:

$\bullet$ \textbf{Stage 1}: We pre-train the first STGNN on GCT flows for feature extraction, separate from the framework's training. This enables the model to focus on the attributes of GCT flows, thus mitigating the impact of \textit{magnitude disparity}.

$\bullet$ \textbf{Stage 2}: 
We transform the features extracted in Stage 1 to align with the amount of mobility flows, addressing the \textit{amount disparity}. The secondary STGNN is then used to process them and predict the mobility flows.

\subsection{Stage 1 of Framework}
\noindent \textbf{Insight.} The first STGNN model on GCT flows, separately from the framework, enhances focus on capturing spatial-temporal patterns, thus yielding enriched features. This distinct training approach simplifies the learning process and reduces the risk of overfitting \cite{lin2023ctcam}.

\noindent \textbf{Notations:}

$\bullet$ $X$: GCT flows of size $[N, D]$, regarded as $N$ road segment with $D$ observations.

$\bullet$ $\mathcal{G}_{gct}$: The graph structure representing connections among road segments collecting GCT flows.

$\bullet$ $h_{i}$: Multi-channel feature of GCT flow $i$, sized $[C, D]$, representing $D$ dimensions across $C$ channels.

$\bullet$ $H$: The set of all $h_{i}$, denoted $H=\{h_{i}\}$, sized $[N, C, D]$.

$\bullet$ $STGNN^{1st}$: The pre-trained STGNN in Stage 1, used for feature extraction.

\noindent \textbf{Implementation:}

\noindent \textbf{\textit{Training.}} We utilize existing STGNNs (e.g., \cite{li2023dynamic,lin2024teltrans}) trained for feature extraction. Following the traffic prediction \cite{wu2019graph}, we train the STGNN to predict \textit{\textbf{N}} GCT flows in the future $D'$ steps, based on $X$ (sized [\textit{\textbf{N}}, $D]$). Details on the data setup are available\footnote{\href{https://github.com/nnzhan/Graph-WaveNet}{https://github.com/nnzhan/Graph-WaveNet}}.

\noindent \textbf{\textit{Extracted Feature.}} 
STGNN models often encode the input $X$ into multi-channel features $H$ (sized $[N, C, D]$) to enrich the representation, with each channel encapsulating distinct spatial-temporal dynamics. Once trained, we regard the output of the STGNN as the \textit{extracted feature}, denoted as:
\begin{equation}
H = STGNN^{1st}(X,\mathcal{G}_{gct}),
\label{eq:extraction}
\end{equation}
where $STGNN^{1st}$ is the pre-trained STGNN in Stage 1, and $H = \{h_1, h_2, \dots, h_N\} \in \mathbb{R}^{N \times C \times D}$, with each $h_i$ representing the multi-channel feature of GCT flow of segment $i$.

\subsection{Stage 2 of Framework}
Stage 2 uses the extracted feature $H$ from Stage 1 to generate mobility flow predictions, comprising \textbf{three steps} as follows:
\subsubsection{Transformation Step}
\noindent \textbf{Insight.} Due to the misalignment between the amounts of GCT and mobility flows, the extracted feature $H$ cannot be directly mapped to individual mobility flows. Thus, we transform $H$ into representations that align with the amounts of mobility flows, integrating directionality within each route.

\noindent \textbf{Notations.}

$\bullet$ ${h}_{\overline{ij}}$: The representation for the mobility flow of route $\overline{ij}$.

$\bullet$ $\overline{H}$: The set of all ${h}_{\overline{ij}}$, as $\overline{H}=\{{h}_{\overline{ij}}\}$, sized $[M, C, D]$.

\noindent \textbf{Implementation.}
To incorporate directionality, we denote $\overline{ij}$ as the result of subtracting the extracted feature $h_{i}$ of the starting segment $i$ from $h_{j}$ of the ending segment $j$, as:
\begin{equation}
{h}_{\overline{ij}} = \sigma({h}_j - {h}_i), 
\label{eq:init_flow}
\end{equation}
where $\sigma(\cdot)$ is a nonlinear function. After process for all $M$ routes, we obtain the initial representation set:
\begin{equation}
\overline{H} = \{h_{\overline{ij}}\}, 
\label{eq:init_flow_all}
\end{equation}
where $\overline{ij}\in \mathbb{R}^{M}$ and $\overline{H}\in \mathbb{R}^{M \times C \times D}$.

\subsubsection{Enhancement Step}

\noindent \textbf{Insight.}
While the derived ${h}_{\overline{ij}}$ corresponds to the mobility flow $\overline{ij}$, it may not capture correlations with neighboring routes, potentially overlooking factors such as congestion propagation from upstream routes \cite{saberi2020simple,yidan2021learning}. Thus, we enrich these representations by integrating interactions among a route's upstream neighbors.

\noindent \textbf{Notations.}

$\bullet$ $\overline{ki}$: The 1-hop upstream neighbor of route $\overline{ij}$, where segment $k$ leads directly into the start segment $i$ of route $\overline{ij}$.

$\bullet$ $\{h_{\overline{ki}}\}$: The set of representations for all 1-hop upstream neighbors of route $\overline{ij}$.

$\bullet$ $\overline{H}_{\overline{ij}}$: The set of representations comprised of route $\overline{ij}$ and its upstream neighbors $\{\overline{ki}\}$.

$\bullet$ $h_{\overline{ij}}^c$: c-th channel representation of the mobility flow $\overline{ij}$.

$\bullet$ ${h}^{'}_{\overline{ij}}$: The enhanced representation of route $\overline{ij}$ after fusion.

$\bullet$ $\overline{H}^{'}$: The set of all ${h}^{'}_{\overline{ij}}$, denoted $\overline{H}^{'}=\{{h}^{'}_{\overline{ij}}\}$.

\noindent \textbf{Implementation.}

\noindent \textit{Preliminary}.
While Graph Attention Networks (GAT) \cite{velivckovic2018graph} are adept at exploring interactions among features and adaptively adjusting weights \cite{zhao2020multivariate}, current GATs fall short in exploring correlations between multi-channel features as they apply uniform weights across all channels. This process may potentially overlook channels that are critical for prediction \cite{brody2022how}.

\noindent \textit{Solution}.
We employ the concept of \textit{\textbf{Multi-Channel GAT (MGAT)}} \cite{lin2023pay}, which is simple but effectively handles multi-channel representations. Below, we briefly outline how we applied MGAT in the fusion process:

\begin{enumerate}
    \item We concatenate each $h_{\overline{ij}}$ with its upstream neighbors $\{h_{\overline{ki}}\}$, as $\overline{H}_{\overline{ij}}$ with size $[Z, C, D]$, $Z = 1 + |\{h_{\overline{ki}}\}|$.
    \item We explore the interactions among entities in $\overline{H}_{\overline{ij}}$. To determine channel-specific weights, MGAT employs $C$ independent GATs, each focusing on the $c$-th channel representation $\overline{H}^{c}_{\overline{ij}} \in \mathbb{R}^{Z \times D}$. 
    \item Specifically, for $c$-th channel representation $\overline{H}^{c}_{\overline{ij}}$, the attention coefficient in \cite{velivckovic2018graph}, denoted as: $e(h^{c}_{\overline{ij}}, h^{c}_{\overline{ki}})$, is employed to calculate the importance of route $\overline{ki}$ to route $\overline{ij}$. These coefficients are normalized by the Softmax function across all neighbors of node $i$, denoted as the attention function: $\alpha^{c}_{\overline{ij}, \overline{ki}} = softmax(e(h^{c}_{\overline{ij}},h^{c}_{\overline{ki}}))$. MGAT then computes a weighted sum of the features for node $i$ and its neighbors, and concatenates results from $C$ independent attention mechanisms: $ {h}^{'}_{\overline{ij}}=\|_{c=1}^{C}(\sigma(\sum_{k }\alpha^{c}_{\overline{ij}, \overline{ki}}{h}^{c}_{\overline{ki}}))$.
\end{enumerate}

We denote the above process (steps 1-3) as
\begin{equation}
{h}^{'}_{\overline{ij}} = MGAT(\overline{H}_{\overline{ij}},\mathcal{G}_{\overline{ij}}),
\label{eq:channel_concat}
\end{equation}
where $\mathcal{G}_{\overline{ij}}$ is the graph structure among routes $\overline{ij}$ and $\overline{ki}$.

After processing all $M$ routes with Equation \ref{eq:channel_concat} to integrate insights from neighbors, we obtain the enhanced sets:
\begin{equation}
\overline{H}^{'}=\{{h}^{'}_{\overline{ij}}\},
\label{eq:channel_concat_all}
\end{equation}
where $\overline{H}^{'} \in \mathbb{R}^{M \times C \times D}$.

\subsubsection{Prediction Step}
\noindent \textbf{Insight.} 
Given that the enhanced representations $\overline{H}^{'}$ align with the amounts of mobility flows, we further apply a secondary STGNN to capture the spatial-temporal dynamics within these representations and generate predictions.

\noindent \textbf{Notations.}

$\bullet$ $STGNN^{2nd}$: The STGNN in Stage 2 for generating mobility flow predictions.

$\bullet$ $\mathcal{G}_{mob}$: The graph structure of connections among routes.

$\bullet$ $Y$: The output of mobility flow prediction, sized $[M, D^{'}]$, representing $M$ mobility flows and $D^{'}$ future steps.

$\bullet$ $MLP$: A multilayer perception, is a fully connected neural network.

\noindent \textbf{Implementation.} 
we employ a secondary STGNN ($STGNN^{2nd}$), denoted as:
\begin{equation}
\hat{H} = STGNN^{2nd}(\overline{H}^{'}, \mathcal{G}_{mob}),
\end{equation}

Following $STGNN^{2nd}$, an MLP is employed to transform $\hat{H}$ into the prediction output format:
\begin{equation}
Y = MLP(\hat{H}).
\end{equation}
Here, the MLP achieves nonlinear transformations to map the high-level features of $STGNN^{2nd}$ to the desired output.

\subsubsection{Framework Training}
We fix the hyperparameters of the pre-trained $STGNN^{1st}$ in Stage 1 to ensure consistency. The feature extracted in Stage 1 is fed forward through Stage 2 to generate mobility flow predictions. We adopt the Mean Absolute Error (MAE) as our loss function, evaluating the accuracy of predictions against the ground truth in our dataset. The MAE is minimized by tuning the hyperparameters of the transformation, enhancement, and prediction steps to achieve optimal accuracy. Details are provided at: 
\href{https://github.com/cy07gn/TeltoMob/tree/main/Model}{https://github.com/cy07gn/TeltoMob/tree/main/Model}

\begin{table*}[ht]
\centering
\small

\begin{tabular}{@{}lccc ccc ccc ccc @{}}
\toprule
 & \multicolumn{3}{c}{\textbf{15 min.}} & \multicolumn{3}{c}{\textbf{30 min.}} & \multicolumn{3}{c}{\textbf{60 min.}} & \multicolumn{3}{c}{\textbf{Overall$^{3}$}} \\
\cmidrule(r){2-4} \cmidrule(lr){5-7} \cmidrule(lr){8-10}  \cmidrule(lr){11-13}  
\textbf{STGNN models$^{1}$} & MAE & RMSE & MAPE & MAE & RMSE & MAPE & MAE & RMSE & MAPE & MAE & RMSE & MAPE\\
\midrule
    DMGCN(w/o$^{1}$) & 3.99 & 7.21 & 39.9\% & 4.05 & 7.38 & 40.7\% & 4.39 & 8.35 & 42.7\% & 
    4.14 & 7.65 & 41.1\%
    \\

    DMGCN(w) & {3.61} & {6.26} & {36.6\%} & {3.65} & {6.31} & {37.1\%} & {3.80} & {6.89} & {38.9\%} 
    & 3.69 & 6.49 & 37.5\%
    \\

    \hdashline
    \textit{IR$^{2}$} 
    & \textbf{9.5\%} & \textbf{13.2\%} & \textbf{8.3\%} & \textbf{9.9\%} & \textbf{14.5\%} & \textbf{8.8\%} & \textbf{13.4\%} & \textbf{17.5\%} & \textbf{8.9\%}
    & \textbf{11.0\%} & \textbf{15.2\%} & \textbf{8.7\%}
    \\

    \midrule

    ESG(w/o) & 3.87 & 6.63 & 39.7\% & 4.01 & 7.28 & 41.1\% & 4.21 & 7.89 & 43.2\% 
    & 4.03 & 7.27 & 41.3\%
    \\
    
    ESG(w) & {3.59} & {6.27} & {37.3\%} & {3.65} & {6.33} & {37.7\%} & {3.76} & {6.56} & {38.4\%} 
    & 3.67 & 6.39 & 37.8\%
    \\

    \hdashline
    \textit{IR} 
    & \textbf{7.2\%} & \textbf{5.4\%} & \textbf{6.0\%} & \textbf{9.0\%} & \textbf{13.1\%} & \textbf{8.3\%} & \textbf{10.7\%} & \textbf{16.9\%} & \textbf{11.1\%} 
    & \textbf{9.0\%} & \textbf{12.1\%} & \textbf{8.5\%}
    \\

     \midrule

     DGCRN(w/o) & 3.86 & 7.09 & 39.4\% & 3.92 & 7.22 & 39.9\% & 4.10 & 7.69 & 42.2\%
     & 3.96 & 7.33 & 40.5\%
     \\
    
    DGCRN(w) & {3.58} & {6.25} & {37.4\%} & {3.61} & {6.30} & {37.6\%} & {3.74} & {6.46} & {38.1\%} 
    & 3.64 & 6.34 & 37.7\%
    \\

    \hdashline
    \textit{IR} 
    & \textbf{7.3\%} & \textbf{11.9\%} & \textbf{5.1\%} & \textbf{7.9\%} & \textbf{12.7\%} & \textbf{5.8\%} & \textbf{8.8\%} & \textbf{16.0\%} & \textbf{9.7\%} 
    & \textbf{8.0\%} & \textbf{13.6\%} & \textbf{6.9\%}
    \\

     \midrule
    MFGM(w/o) & 3.72 & 6.41 & 38.3\% & 3.84 & 6.66 & 38.9\% & 4.01 & 7.41 & 40.6\% 
    & 3.86 & 6.83 & 39.27\%
    \\
    
    MFGM(w) & 3.45 & 5.69 & 34.7\% & 3.54 & 5.89 & 34.9\% & 3.69 & 6.41 & 36.2\% 
    & 3.56 & 6.00 & 35.29\%
    \\

    \hdashline
    \textit{IR} 
    & \textbf{7.3\%} & \textbf{11.2\%} & \textbf{9.4\%} & \textbf{7.8\%} & \textbf{11.6\%} & \textbf{10.2\%} & \textbf{8.0\%} & \textbf{13.5\%} & \textbf{10.7\%} 
    & \textbf{7.68\%} & \textbf{12.10\%} & \textbf{10.11\%}
    \\

    \midrule
    Average IR 
    & \textbf{7.8\%} & \textbf{10.4\%} & \textbf{7.2\%}
    & \textbf{8.6\%} & \textbf{13.0\%} & \textbf{8.3\%}
    & \textbf{10.2\%} & \textbf{16.0\%} & \textbf{10.1\%} 
    & \textbf{8.9\%} & \textbf{13.2\%} & \textbf{8.6\%} 
    \\

\bottomrule
\end{tabular}
\scriptsize
\begin{tablenotes}\scriptsize
\item $^{1}$The model is used for prediction without (w/o) integration into our framework.
\item $^{2}$IR (Improvement Ratio) = ((score(w/0) - score(w)) / score(w/o)) * 100\%.
\item $^{3}$Average results from 15 min to 60 min.
\end{tablenotes}
 \caption{Performance Comparisons With/Without Framework.}
\label{pred_table}
\end{table*}

\section{Experiments}
\subsection{Experimental Setup}
\label{Experimental_Setup}
\noindent \textbf{Data Setups.}
We collected data at 15-minute intervals from 2022/8/28 to 2022/9/27, yielding 2,976 samples of GCT and mobility flows across 34 road segments and 84 routes. Sequences for the Train/Test/Valid were formed from these samples, each comprising 12 steps: the initial 8 steps ($T_{in}$) as historical GCT flows and the next 4 steps ($T_{out}$) as future mobility flows. Following \cite{li2018diffusion}, we divided these sequences into Train/Test/Valid sets in a 70\%-20\%-10\% ratio. Each experiment runs for 180 epochs with early stopping.

\noindent \textbf{Baselines.}
We chosen representative STGNN baselines integrated into our framework for this new task: {\textbf{{DMGCN}}} \cite{han2021dynamic}: Leverages time-specific spatial dependencies with a multi-faceted fusion. {\textbf{{ESG}}}\cite{ye2022learning}: Employs evolutionary and multi-scale graph structures. \textbf{{{DGCRN}}} \cite{li2023dynamic}: Models the dynamic graph with a seq2seq architecture. {\textbf{{MFGM}}}\cite{lin2024teltrans}: Captures multivariate, temporal, and spatial dynamics with a GNN-based approach.

\noindent \textbf{Metrics.}
We use Mean Absolute Error (MAE), Root Mean Squared Error (RMSE), and Mean Absolute Percentage Error (MAPE) to assess our predictions against ground truth mobility flows from 15 (1 step) to 60 minutes (4 steps).

\subsection{Prediction Performance}
\label{prediction}
\noindent \textbf{Performance Improvement with Framework.}
We evaluated the integration of various STGNN models with our framework, focusing on prediction intervals ranging from 15 to 60 minutes. Each model was examined under two settings: without the framework integration (\textbf{w/o}) and with our framework integration (\textbf{w}). For the \textbf{w/o} setting, we used the STGNN, inputting the GCT flow from each route's starting segment, to output the predicted mobility flows.

Table \ref{pred_table} presents the performance of all models in both settings, with each reported result representing the average of 10 individual runs. We use the Improvement Ratio (\textbf{IR}) to measure the enhancement achieved by integrating STGNN models into our framework. The results reveal that this integration boosts performance, with overall average IRs of 8.9\%, 13.2\%, and 8.6\% for MAE, RMSE, and MAPE, respectively, and up to a 17.5\% RMSE improvement for long-term predictions. This underlines the compatibility of our framework across different STGNN models and its effectiveness.

Notably, as the prediction interval lengthens, performance typically declines due to the increased complexity of long-term dependencies. However, models enhanced with our framework consistently improve in prediction accuracy, achieving progressively larger IRs as the forecast duration extends. Specifically, the average IR for MAE, RMSE, and MAPE grew from 7.8\%, 10.4\%, and 7.2\% at 15 minutes to 10.2\%, 16.0\%, and 10.1\% at 60 minutes, respectively. These findings underscore our framework's capability for more complex, long-term predictions, which is practical for real-world applications \cite{tian2021spatial}.

\noindent \textbf{Computational Efficiency}
Table \ref{reaction_time} presents the computational efficiency of various STGNN models within our framework on a Nvidia Tesla T4 GPU, with each value representing the average of 10 runs. DMGCN and MFGM show promising inference times (0.73 and 0.79 seconds respectively), suitable for near real-time applications, while ESG and DGCRN are slightly slower. Regarding training times, MFGM is most efficient at 13.53 seconds, suggesting an advantage in environments requiring rapid model updates, whereas DMGCN and ESG were slower, which might impact their adaptability in environments with rapidly changing data. This assessment indicates that our framework is capable of providing efficient inference, supporting its potential for integration within real-time transportation systems, as depicted in Section \ref{applications}.

\begin{table}[h]
    \centering
    \small
    \begin{tabular}{lll}
        \hline
        \textbf{STGNN models} & Inference Time & Training Time \\
        \hline
           DMGCN & 0.73 sec & 16.58 secs/epoch \\
    ESG   & 1.65 secs & 15.93 secs/epoch \\
    DGCRN & 1.58 secs & 14.56 secs/epoch \\
    {MFGM}& 0.79 secs & 13.53 secs/epoch \\
        \hline
    \end{tabular}
    \caption{Comparisons for Inference and Training Times.}
    \label{reaction_time}
\end{table}

\subsection{Ablation Study of Our Framework}
\label{Ablation_Study}
We assessed the contributions of framework's components by comparing the framework with \textbf{\textit{four ablated settings}}: without the Pre-trained STGNN (\textbf{w/o STGNN$^{1st}$}), without the Transformation step (\textbf{w/o Trans.}), without the Enhancement step (\textbf{w/o Enhan.}), and without Stage 2's STGNN (\textbf{w/o STGNN$^{2nd}$}). Table \ref{table:ablation} shows the average results for prediction length 15 min to 60 min, ordered by \textit{performance impact}:
 \begin{table}[h]
\begin{threeparttable}
    \scriptsize
    
    \setlength\tabcolsep{0pt}
\begin{tabular*}{\linewidth }{@{\extracolsep{\fill}} l c c c | c c c | c c c  @{} }
    \toprule

    &   & 15min. &  
       &  & 30min. &  &  & 60min. &
         \\

   \textbf{Ablation models}     & MAE & RMSE & MAPE 
         & MAE & RMSE & MAPE 
         & MAE & RMSE & MAPE 
\\
\midrule
   w/o STGNN$^{1st}$

& 3.77 & 7.18 & 37.2\% & 3.91 & 7.41 & 38.2\% & 4.42 & 8.61 & 40.1\% \\

w/o Enhan.
& 3.65 & 6.53 & 36.3\% & 3.78 & 6.87 & 36.9\% & 4.13 & 7.69 & 39.4\% \\

w/o STGNN$^{2nd}$
& 3.58 & 6.27 & 35.9\% & 3.67 & 6.43 & 36.4\% & 3.97 & 7.32 & 37.9\% \\

    w/o Trans.
& 3.52 & 6.09 & 35.4\% & 3.62 & 6.16 & 35.6\% & 3.82 & 6.94 & 37.5\% \\

    \textbf{Full Framework}
& \textbf{3.45} & \textbf{5.69} & \textbf{34.7\%} & \textbf{3.54} & \textbf{5.89} & \textbf{34.9\%} & \textbf{3.69} & \textbf{6.41} & \textbf{36.2\%}\\

        \bottomrule

\end{tabular*}

\caption{Ablation Study of Our Framework.}
\label{table:ablation}
\end{threeparttable}
\end{table}

\noindent \textbf{Impact of w/o STGNN$^{1st}$.} 
This setting omits the pre-trained STGNN$^{1st}$ from Stage 1, using raw GCT flows instead of the extracted features that capture spatio-temporal dynamics. Without these extracted implicit features within the GCT flow, this configuration demonstrates the worst performance metrics across all intervals, indicating a significant decrease in accuracy. This suggests that the pre-trained STGNN to capture the underlying patterns in GCT flows is very crucial.

\noindent \textbf{Impact of w/o Enhan.} 
This setting excludes the Enhancement step, thereby omitting the incorporation of correlations between each route and its upstream neighbors. This omission leads to the second-largest performance decrease. We argue that, given the spatial dependencies among routes as shown in our dataset (see Figure \ref{radar}), overlooking these correlations might miss crucial insights, such as congestion propagation from upstream, thus decreasing the performance.

\noindent \textbf{Impact of w/o STGNN$^{2nd}$.} 
This setting omits STGNN$^{2nd}$ in the Prediction step, opting for MLPs to generate the predictions. Although this removal is not as severe as omitting the STGNN$^{1st}$, it still consistently increases prediction errors across all intervals. This validates that capturing implicit dynamics with STGNN$^{2nd}$ contributes to the outcomes.

\noindent \textbf{Impact of w/o Trans.}
This setting omits the Transformation step, directly using the extracted GCT flow feature as a representation of mobility flow, without integrating the directionality among routes. Although excluding the transformation step leads to slightly worse metrics, it still leads to increased errors across the 15 to 60 minutes, confirming that incorporating directionality can enhance mobility flow prediction.

\begin{figure}[ht]
\centering
\includegraphics[width=0.95\linewidth]{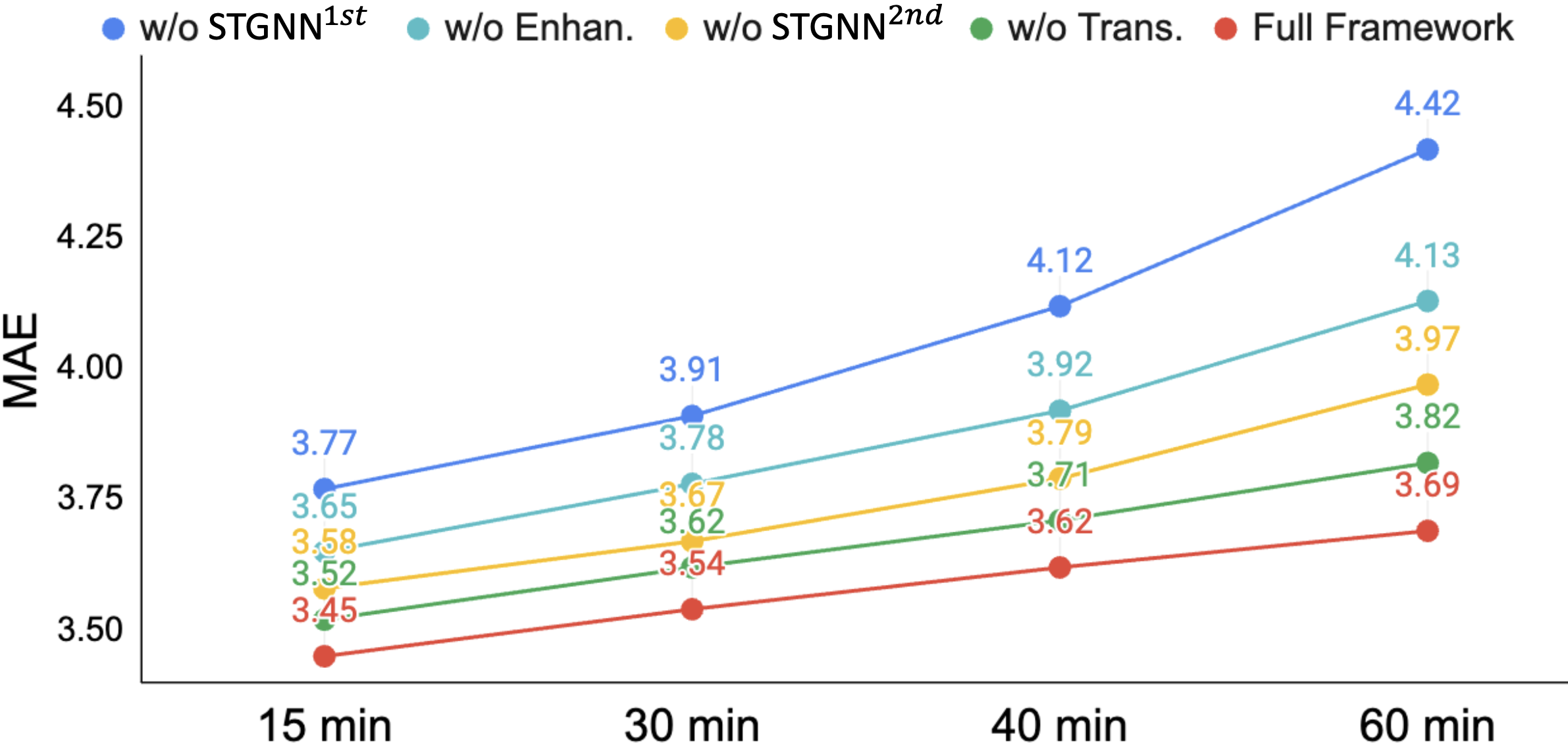}
\caption{Ablation study of MAE across 15 to 60-minute intervals. As prediction lengths extend, performance declines for all settings, while the full framework (red line) not only consistently outperforms ablated versions but with a growing MAE gap against them, proving the essentiality of all components.}
\label{fig:mae}
\end{figure}

Figure \ref{fig:mae} presents the predictive performance as measured by MAE across time intervals. As the interval lengthens, the error for all settings increases. However, it is observed that the full framework consistently outperforms the other ablated settings at all prediction lengths, with the MAE gap widening over time. This not only demonstrates the superior performance of the full framework but also highlights its stability for complex, long-term tasks.
\subsection{Path Forward for Real-World Deployment}
\label{applications}
\begin{figure}[ht]
\centering
\includegraphics[width=0.95\linewidth]{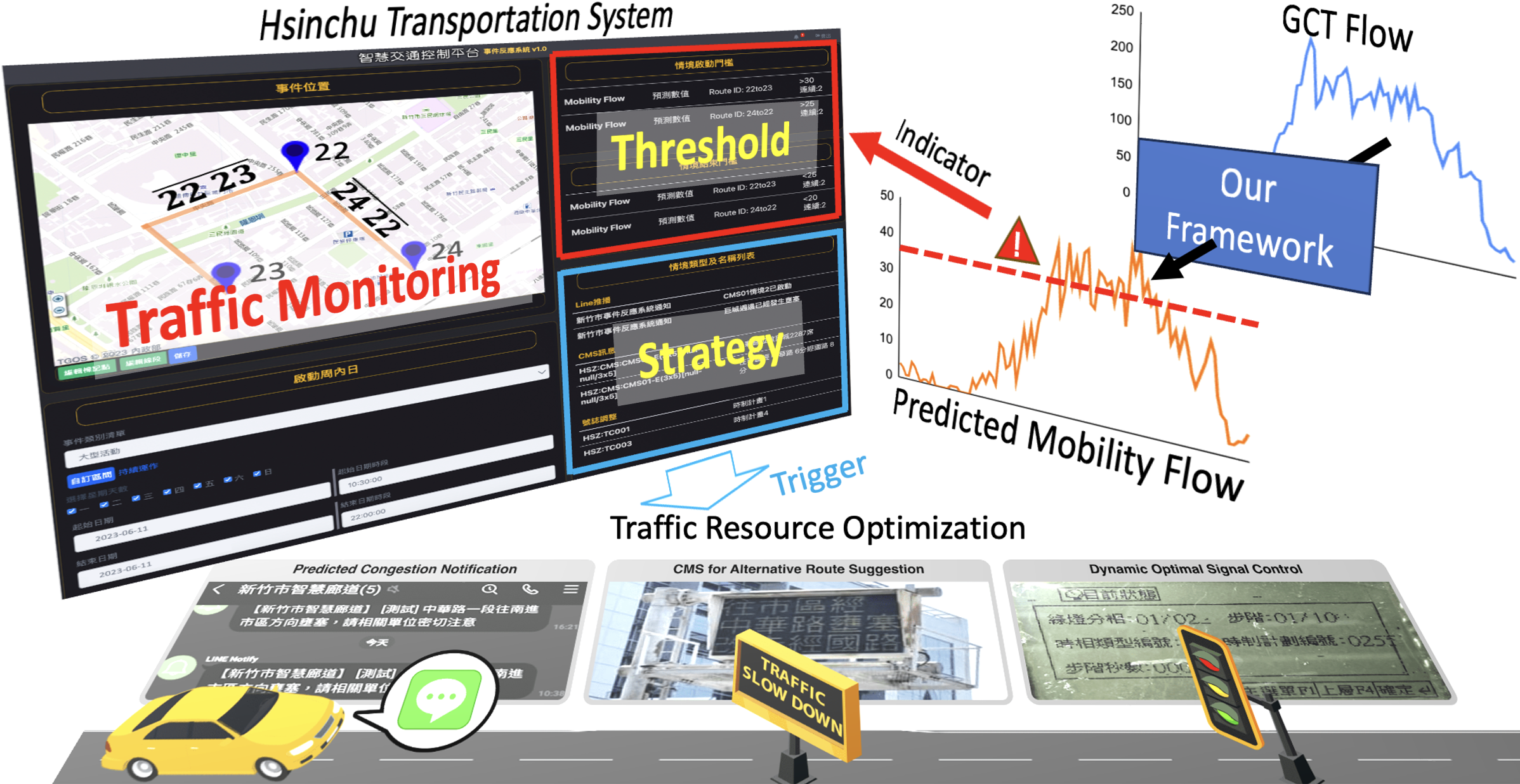}
\caption{Integrating our framework with the city's transportation system, by utilizing GCT flow for predictive mobility insights, activates a threshold-based alert system for optimal traffic management. This achieves practical convergence of telecom data and transportation needs through driver notifications, CMS for alternative routing, and optimized traffic signal control.}
\label{fig:applications}
\end{figure}

Our framework utilizing GCT flow to forecast mobility flow shows promising potential, while with efficient inference (as Table \ref{reaction_time}). As depicted in Figure \ref{fig:applications}, we are collaborating with city authorities to integrate this framework into the transportation system, focusing on \textit{two keys}:

$\bullet$ \textbf{Traffic Monitoring}: Predicted mobility flows offer insights for authorities to monitor potential congestion.

$\bullet$ \textbf{Traffic Indicator}: The system employs these forecasts in a threshold-based alert mechanism, serving as a new indicator of traffic conditions. When pre-set thresholds are exceeded, it triggers various strategies: sending notifications to authorities, suggesting alternative routes through Changeable Message Signs (CMS) to redirect commuters, and dynamically adjusting traffic signal plans to optimize flow and delays.

These applications highlight the potential of integrating directional telecom network-generated data into transportation systems. We anticipate scaling our Proof-of-Concept to broader areas, thus reducing the reliance on extensive sensor deployments and advancing sustainable urban environments.

\section{Conclusion}
We leverage undirected \textbf{telecom data} to forecast directional mobility flows along routes, enhancing the utility of telecom data in transportation and reducing the deployment and maintenance costs of detectors, thus advancing sustainable cities (\textbf{SDG 11}). To tackle the challenge, we propose a two-stage STGNN \textbf{framework}, facilitated by our \textbf{TeltoMob} dataset, to assess its effectiveness. Our experiments confirm the framework's \textbf{compatibility} with various STGNN models and its effectiveness in enhancing their performance, with up to a \textbf{17.5\%} improvement in long-term prediction. We also demonstrate the integration of the framework into the \textbf{transportation system} as a traffic indicator. This work underscores the potential of telecom data in transportation and contributes to the enhancement of sustainable urban mobility.
\clearpage

\section{Acknowledgments}
This work was supported in part by National Science and Technology Council, Taiwan, under Grant NSTC 111-2634-F-002-022 and by Qualcomm through a Taiwan University Research Collaboration Project.

\bibliographystyle{named}
\bibliography{ijcai24}

\begin{thebibliography}{}

\bibitem[\protect\citeauthoryear{Babu and Manoj}{2020}]{babu2020toward}
Sarath Babu and BS~Manoj.
\newblock Toward a type-based analysis of road networks.
\newblock {\em ACM Transactions on Spatial Algorithms and Systems (TSAS)}, 6(4):1--45, 2020.

\bibitem[\protect\citeauthoryear{Brody \bgroup \em et al.\egroup }{2022}]{brody2022how}
Shaked Brody, Uri Alon, and Eran Yahav.
\newblock How attentive are graph attention networks?
\newblock In {\em Proc. of ICLR}, 2022.

\bibitem[\protect\citeauthoryear{Cisco}{2021}]{cisco2021cisco}
U~Cisco.
\newblock Cisco annual internet report (2018--2023) white paper. 2020.
\newblock {\em Acessado em}, 10(01), 2021.

\bibitem[\protect\citeauthoryear{Cohen \bgroup \em et al.\egroup }{2009}]{cohen2009pearson}
Israel Cohen, Yiteng Huang, Jingdong Chen, Jacob Benesty, Jacob Benesty, Jingdong Chen, Yiteng Huang, and Israel Cohen.
\newblock Pearson correlation coefficient.
\newblock {\em Noise reduction in speech processing}, 2009.

\bibitem[\protect\citeauthoryear{Guo \bgroup \em et al.\egroup }{2019}]{guo2019attention}
Shengnan Guo, Youfang Lin, Ning Feng, Chao Song, and Huaiyu Wan.
\newblock Attention based spatial-temporal graph convolutional networks for traffic flow forecasting.
\newblock In {\em Proc. of AAAI}, 2019.

\bibitem[\protect\citeauthoryear{Han \bgroup \em et al.\egroup }{2021}]{han2021dynamic}
Liangzhe Han, Bowen Du, Leilei Sun, Yanjie Fu, Yisheng Lv, and Hui Xiong.
\newblock Dynamic and multi-faceted spatio-temporal deep learning for traffic speed forecasting.
\newblock In {\em Proc. of KDD}, 2021.

\bibitem[\protect\citeauthoryear{He \bgroup \em et al.\egroup }{2020}]{he2020semi}
Qing He, Gy{\"o}rgy D{\'a}n, and Georgios~P Koudouridis.
\newblock Semi-persistent scheduling for 5g downlink based on short-term traffic prediction.
\newblock In {\em GLOBECOM 2020-2020 IEEE Global Communications Conference}, pages 1--6. IEEE, 2020.

\bibitem[\protect\citeauthoryear{Jiang}{2022}]{jiang2022cellular}
Weiwei Jiang.
\newblock Cellular traffic prediction with machine learning: A survey.
\newblock {\em Expert Systems with Applications}, page 117163, 2022.

\bibitem[\protect\citeauthoryear{Li \bgroup \em et al.\egroup }{2018}]{li2018diffusion}
Yaguang Li, Rose Yu, Cyrus Shahabi, and Yan Liu.
\newblock Diffusion convolutional recurrent neural network: Data-driven traffic forecasting.
\newblock In {\em Proc. of ICLR}, 2018.

\bibitem[\protect\citeauthoryear{Li \bgroup \em et al.\egroup }{2023}]{li2023dynamic}
Fuxian Li, Jie Feng, Huan Yan, Guangyin Jin, Fan Yang, Funing Sun, Depeng Jin, and Yong Li.
\newblock Dynamic graph convolutional recurrent network for traffic prediction: Benchmark and solution.
\newblock {\em ACM Transactions on Knowledge Discovery from Data}, 17(1):1--21, 2023.

\bibitem[\protect\citeauthoryear{Lin \bgroup \em et al.\egroup }{2021a}]{lin2021multivariate}
Chung-Yi Lin, Hung-Ting Su, Shen-Lung Tung, and Winston~H. Hsu.
\newblock Multivariate and propagation graph attention network for spatial-temporal prediction with outdoor cellular traffic.
\newblock In {\em Proc. of CIKM}, 2021.

\bibitem[\protect\citeauthoryear{Lin \bgroup \em et al.\egroup }{2021b}]{lin2021data}
Jiansheng Lin, Youjia Chen, Haifeng Zheng, Ming Ding, Peng Cheng, and Lajos Hanzo.
\newblock A data-driven base station sleeping strategy based on traffic prediction.
\newblock {\em IEEE Transactions on Network Science and Engineering}, 2021.

\bibitem[\protect\citeauthoryear{Lin \bgroup \em et al.\egroup }{2023a}]{lin2023pay}
ChungYi Lin, Shen-Lung Tung, and Winston~H Hsu.
\newblock Pay attention to multi-channel for improving graph neural networks.
\newblock In {\em Proc. of ICLR}, 2023.
\newblock Accepted paper, to appear.

\bibitem[\protect\citeauthoryear{Lin \bgroup \em et al.\egroup }{2023b}]{lin2023ctcam}
ChungYi Lin, Shen-Lung Tung, Hung-Ting Su, and Winston~H Hsu.
\newblock Ctcam: Enhancing transportation evaluation through fusion of cellular traffic and camera-based vehicle flows.
\newblock In {\em Proc. of CIKM}, 2023.

\bibitem[\protect\citeauthoryear{Lin \bgroup \em et al.\egroup }{2024}]{lin2024teltrans}
ChungYi Lin, Shen-Lung Tung, Hung-Ting Su, and Winston~H Hsu.
\newblock Teltrans: Applying multi-type telecom data to transportation evaluation and prediction via multifaceted graph modeling.
\newblock {\em arXiv preprint arXiv:2401.03138}, 2024.
\newblock Accepted by AAAI 2024. To appear.

\bibitem[\protect\citeauthoryear{Lv \bgroup \em et al.\egroup }{2021}]{lv2021deep}
Zhihan Lv, Yuxi Li, Hailin Feng, and Haibin Lv.
\newblock Deep learning for security in digital twins of cooperative intelligent transportation systems.
\newblock {\em IEEE Transactions on Intelligent Transportation Systems}, 2021.

\bibitem[\protect\citeauthoryear{Peng \bgroup \em et al.\egroup }{2016}]{peng2016computational}
Chi-Han Peng, Yong-Liang Yang, Fan Bao, Daniel Fink, Dong-Ming Yan, Peter Wonka, and Niloy~J Mitra.
\newblock Computational network design from functional specifications.
\newblock {\em ACM Transactions on Graphics (TOG)}, 35(4):1--12, 2016.

\bibitem[\protect\citeauthoryear{Saberi \bgroup \em et al.\egroup }{2020}]{saberi2020simple}
Meead Saberi, Homayoun Hamedmoghadam, Mudabber Ashfaq, Seyed~Amir Hosseini, Ziyuan Gu, Sajjad Shafiei, Divya~J Nair, Vinayak Dixit, Lauren Gardner, S~Travis Waller, et~al.
\newblock A simple contagion process describes spreading of traffic jams in urban networks.
\newblock {\em Nature communications}, 2020.

\bibitem[\protect\citeauthoryear{Sen \bgroup \em et al.\egroup }{2012}]{sen2012kyun}
Rijurekha Sen, Abhinav Maurya, Bhaskaran Raman, Rupesh Mehta, Ramakrishnan Kalyanaraman, Nagamanoj Vankadhara, Swaroop Roy, and Prashima Sharma.
\newblock Kyun queue: a sensor network system to monitor road traffic queues.
\newblock In {\em Proc. of SENSYS}, pages 127--140, 2012.

\bibitem[\protect\citeauthoryear{Tian and Chan}{2021}]{tian2021spatial}
Chenyu Tian and Wai~Kin Chan.
\newblock Spatial-temporal attention wavenet: A deep learning framework for traffic prediction considering spatial-temporal dependencies.
\newblock {\em IET Intelligent Transport Systems}, 2021.

\bibitem[\protect\citeauthoryear{Veli{\v{c}}kovi{\'c} \bgroup \em et al.\egroup }{2018}]{velivckovic2018graph}
Petar Veli{\v{c}}kovi{\'c}, Guillem Cucurull, Arantxa Casanova, Adriana Romero, Pietro Lio, and Yoshua Bengio.
\newblock Graph attention networks.
\newblock In {\em Proc. of ICLR}, 2018.

\bibitem[\protect\citeauthoryear{Wang \bgroup \em et al.\egroup }{2018}]{wang2018spatio}
Xu~Wang, Zimu Zhou, Fu~Xiao, Kai Xing, Zheng Yang, Yunhao Liu, and Chunyi Peng.
\newblock Spatio-temporal analysis and prediction of cellular traffic in metropolis.
\newblock {\em IEEE Transactions on Mobile Computing}, 2018.

\bibitem[\protect\citeauthoryear{Wang \bgroup \em et al.\egroup }{2022}]{wang2022spatial}
Zi~Wang, Jia Hu, Geyong Min, Zhiwei Zhao, Zheng Chang, and Zhe Wang.
\newblock Spatial-temporal cellular traffic prediction for 5 g and beyond: A graph neural networks-based approach.
\newblock {\em IEEE Transactions on Industrial Informatics}, 2022.

\bibitem[\protect\citeauthoryear{Wu \bgroup \em et al.\egroup }{2019}]{wu2019graph}
Zonghan Wu, Shirui Pan, Guodong Long, Jing Jiang, and Chengqi Zhang.
\newblock Graph wavenet for deep spatial-temporal graph modeling.
\newblock In {\em Proc. of IJCAI}, 2019.

\bibitem[\protect\citeauthoryear{Xie \bgroup \em et al.\egroup }{2020}]{xie2020urban}
Peng Xie, Tianrui Li, Jia Liu, Shengdong Du, Xin Yang, and Junbo Zhang.
\newblock Urban flow prediction from spatiotemporal data using machine learning: A survey.
\newblock {\em Information Fusion}, 2020.

\bibitem[\protect\citeauthoryear{Yao \bgroup \em et al.\egroup }{2021}]{yao2021mvstgn}
Yang Yao, Bo~Gu, Zhou Su, and Mohsen Guizani.
\newblock Mvstgn: A multi-view spatial-temporal graph network for cellular traffic prediction.
\newblock {\em IEEE Transactions on Mobile Computing}, 2021.

\bibitem[\protect\citeauthoryear{Ye \bgroup \em et al.\egroup }{2022}]{ye2022learning}
Junchen Ye, Zihan Liu, Bowen Du, Leilei Sun, Weimiao Li, Yanjie Fu, and Hui Xiong.
\newblock Learning the evolutionary and multi-scale graph structure for multivariate time series forecasting.
\newblock In {\em Proc. of KDD}, pages 2296--2306, 2022.

\bibitem[\protect\citeauthoryear{Yidan \bgroup \em et al.\egroup }{2021}]{yidan2021learning}
Sun Yidan, He~Peilan, Jiang Guiyuan, and Siew-Kei Lam.
\newblock Learning congestion propagation behaviors for traffic prediction.
\newblock In {\em 2021 IEEE International Intelligent Transportation Systems Conference (ITSC)}, pages 2175--2180. IEEE, 2021.

\bibitem[\protect\citeauthoryear{Zhang \bgroup \em et al.\egroup }{2018}]{zhang2018citywide}
Chuanting Zhang, Haixia Zhang, Dongfeng Yuan, and Minggao Zhang.
\newblock Citywide cellular traffic prediction based on densely connected convolutional neural networks.
\newblock {\em IEEE Communications Letters}, 2018.

\bibitem[\protect\citeauthoryear{Zhao \bgroup \em et al.\egroup }{2020}]{zhao2020multivariate}
Hang Zhao, Yujing Wang, Juanyong Duan, Congrui Huang, Defu Cao, Yunhai Tong, Bixiong Xu, Jing Bai, Jie Tong, and Qi~Zhang.
\newblock Multivariate time-series anomaly detection via graph attention network.
\newblock In {\em Proc. of ICDM}, pages 841--850, 2020.

\bibitem[\protect\citeauthoryear{Zhao \bgroup \em et al.\egroup }{2021}]{zhao2021spatial}
Nan Zhao, Aonan Wu, Yiyang Pei, Ying-Chang Liang, and Dusit Niyato.
\newblock Spatial-temporal aggregation graph convolution network for efficient mobile cellular traffic prediction.
\newblock {\em IEEE Communications Letters}, 2021.

\end{thebibliography}

\end{document}